\documentclass[sigconf]{acmart}

\usepackage{amsmath,amssymb,amsfonts}
\usepackage{algorithmic}
\usepackage{graphicx}
\usepackage{textcomp}
\usepackage{xcolor}
\usepackage[ruled,lined,linesnumbered,vlined,algo2e]{algorithm2e}
\usepackage{url}
\usepackage{multirow}
\usepackage{array}
\usepackage{booktabs}
\usepackage{multicol}
\usepackage{makecell}
\usepackage{subfig}
\usepackage[font=small]{caption}
\usepackage{threeparttable}
\usepackage{verbatim}
\usepackage{color}
\usepackage{tcolorbox}
\definecolor{fan}{rgb}{255,0,0} 
\newcommand{\mytool}{ExpGA}
\AtBeginDocument{%
  \providecommand\BibTeX{{%
    \normalfont B\kern-0.5em{\scshape i\kern-0.25em b}\kern-0.8em\TeX}}}

\copyrightyear{2022}
\acmYear{2022}
\setcopyright{acmcopyright}\acmConference[ICSE '22]{44th International Conference
	on Software Engineering}{May 21--29, 2022}{Pittsburgh, PA, USA}
\acmBooktitle{44th International Conference on Software Engineering (ICSE '22), May
	21--29, 2022, Pittsburgh, PA, USA}
\acmPrice{15.00}
\acmDOI{10.1145/3510003.3510137}
\acmISBN{978-1-4503-9221-1/22/05}

\begin{document}

\title{Explanation-Guided Fairness Testing through Genetic Algorithm}

\author{Ming Fan}
\email{mingfan@mail.xjtu.edu.cn}
\affiliation{%
  \institution{Xi'an Jiaotong University, China}
}

\author{Wenying Wei}
\email{waving@stu.xjtu.edu.cn}
\affiliation{%
  \institution{Xi'an Jiaotong University, China}
}

\author{Wuxia Jin}
\email{jinwuxia@mail.xjtu.edu.cn}
\affiliation{%
	\institution{Xi'an Jiaotong University, China}
}

\author{Zijiang Yang}
\email{zijiang@xjtu.edu.cn}
\affiliation{%
	\institution{Xi'an Jiaotong University, China}
}

\author{Ting Liu}
\authornote{Corresponding author.}
\email{tingliu@mail.xjtu.edu.cn}
\affiliation{%
	\institution{Xi'an Jiaotong University, China}
    }


\renewcommand{\shortauthors}{Ming Fan and Wenying Wei, et al.}

\begin{abstract}
The fairness characteristic is a critical attribute of trusted AI systems. A plethora of research has proposed diverse methods for individual fairness testing. 
However, they are suffering from three major limitations, i.e., low efficiency, low effectiveness, and model-specificity. This work proposes ExpGA, an explanation-guided fairness testing approach through a genetic algorithm (GA). \mytool{} employs the explanation results generated by interpretable methods to collect high-quality initial seeds, which are prone to derive discriminatory samples by slightly modifying  feature values.  \mytool{} then adopts GA  to search discriminatory sample candidates by optimizing a fitness value. Benefiting from this combination of explanation results and GA, \mytool{} is both efficient and effective to detect discriminatory individuals. Moreover, \mytool{} only requires prediction probabilities of the tested model, resulting in a better generalization capability to various models. Experiments on multiple real-world benchmarks, including tabular and text datasets, show that \mytool{} presents higher efficiency and effectiveness than four state-of-the-art approaches.
\end{abstract}



\maketitle

\section{Introduction}
\label{sec_introduction}

Artificial Intelligence (AI) systems have been penetrating various aspects of our lives. Except for the benefits brought by AI systems, however, an AI system would harm the fairness of our society if it makes biased decisions and thus presents a discriminatory nature. Recent studies~\cite{buolamwini2018gender,tatman2017gender} pointed out that some familiar AI systems behave with deviations in gender and skin color. For example, in face recognition systems developed by IBM, Microsoft, and face++, the detection accuracy for white-skin men is much higher (sometimes 20\% higher) than that of black-skin women.


Fairness characteristic is critical for trusted AI systems, and fairness testing has attracted much attention recently in both industrial (e.g., IBM's AI Fairness 360~\cite{AI360}, Google's ML-Fairness-Gym~\cite{FairnessGym}) and academic communities~\cite{mehrabi2019survey,saxena2019perceptions,tramer2017fairtest}. The GDPR~\cite{GDPR}, a data privacy law across Europe, claims that ``the personal data shall be processed fairly in relation to the data subject''. Definitions of fairness in existing work fall into two categories~\cite{mehrabi2019survey}: 1) individual fairness, i.e., similar individuals are given similar decisions; 2) group fairness, i.e., equal decisions are made for different groups. Most research on software fairness, including this work, has focused on individual fairness~\cite{brun2018software}.

\textbf{Individual fairness testing} mutates the input samples to generate sufficient testing samples and checks whether these samples are individual-discriminatory. Typical approaches include THEMIS~\cite{galhotra2017fairness}, AEQUITAS~\cite{udeshi2018automated}, SG~\cite{aggarwal2019black}, ADF~\cite{zhang2020white}, and MT-NLP~\cite{ma2020metamorphic}. In general, these fairness testing methods first construct a set of seed samples that are suspiciously discriminatory using random or clustering techniques. Then, they search for more discriminatory samples around the seeds through different strategies, such as symbolic execution and gradient-based methods. After that, they expand the original dataset by modifying the labels of detected discriminatory samples and finally retrain to improve the model fairness.
Prior individual fairness testing approaches have achieved relatively good performance; however, they still suffer from three major limitations.

\noindent{\textbf{Low efficiency}}: When handling complex deep learning systems, SG~\cite{aggarwal2019black} is inefficient due to the use of symbolic execution. Symbolic execution, which typically requires SMT solvers, is computationally expensive to generate particular test cases.

\noindent{\textbf{Low effectiveness}}: THEMIS~\cite{galhotra2017fairness}, AEQUITAS~\cite{udeshi2018automated}, and MT-NLP~\cite{ma2020metamorphic} adopt unguided or semi-guided search strategies, leading to a low success rate of the discriminatory sample generation. 

\noindent{\textbf{Model-specificity}}: ADF~\cite{zhang2020white}, a lightweight approach, requires  gradient information. Such information is extracted from the white-box DNN model but  unavailable in black-box models. 

To address these limitations, this work proposes \mytool{}, a model-agnostic fairness testing approach. \mytool{} employs explanation results and the genetic algorithm (GA), thus detecting discriminatory samples both efficiently and effectively.

Our \mytool{} first leverages the explanation results generated by interpretable methods to identify a set of seed samples. Seed samples are prone to derive discriminatory samples by slightly modifying their feature values. Then, using seed samples as the input, \mytool{} employs the GA to produce large amount of discriminatory offspring. In \mytool{}, the explanation results favor  high-quality seeds.  GA can guide the search process towards discriminatory sample solutions by optimizing the fitness value.  Furthermore, \mytool{} is able to handle black-box models since it only requires prediction probabilities from the models. 

To evaluate the efficiency and effectiveness of our \mytool{}, we compare \mytool{} with four state-of-the-art discriminatory testing approaches, including AEQUITAS, SG, ADF, and MT-NLP. Using five diverse training models, we conduct comprehensive experiments on three tabular dataset benchmarks (handled by AEQUITAS, SG, and ADF)  and two text dataset benchmarks (handled by MT-NLP). The data of our experiments can be found on github\footnote{\url{https://github.com/waving7799/ExpGA}}. 

The results indicate that \mytool{} can detect discriminatory samples more efficiently and effectively than the baseline approaches. In particular, on tabular datasets, \mytool{} consumes less than 0.2s to identify a discriminatory sample with a success rate of 49\%  on average, outperforming AEQUITAS, SG, and ADF in both efficiency and accuracy. On text datasets, \mytool{} is five times more efficient and twice more effective than MT-NLP. In addition, the performance of \mytool{} is more stable in various models than the four baseline approaches, indicating a better generalization capability of \mytool{}.  

In summary, our work makes the following contributions:

\renewcommand\theenumi{\roman{enumi}}
\renewcommand\labelenumi{(\theenumi)}
\begin{enumerate}
	\item {We propose a model-agnostic fairness testing approach called \mytool{}. It can identify individual discriminatory samples for both tabular data and text data by combining the local explanation and GA.}
	\item {We conduct extensive experiments on three tabular datasets and two text datasets to evaluate \mytool{}. The results show that \mytool{} outperforms four state-of-the-art approaches in  terms of both efficiency and effectiveness.}
\end{enumerate}

\begin{figure}[!t]
	\centering
	\includegraphics[scale=0.48]{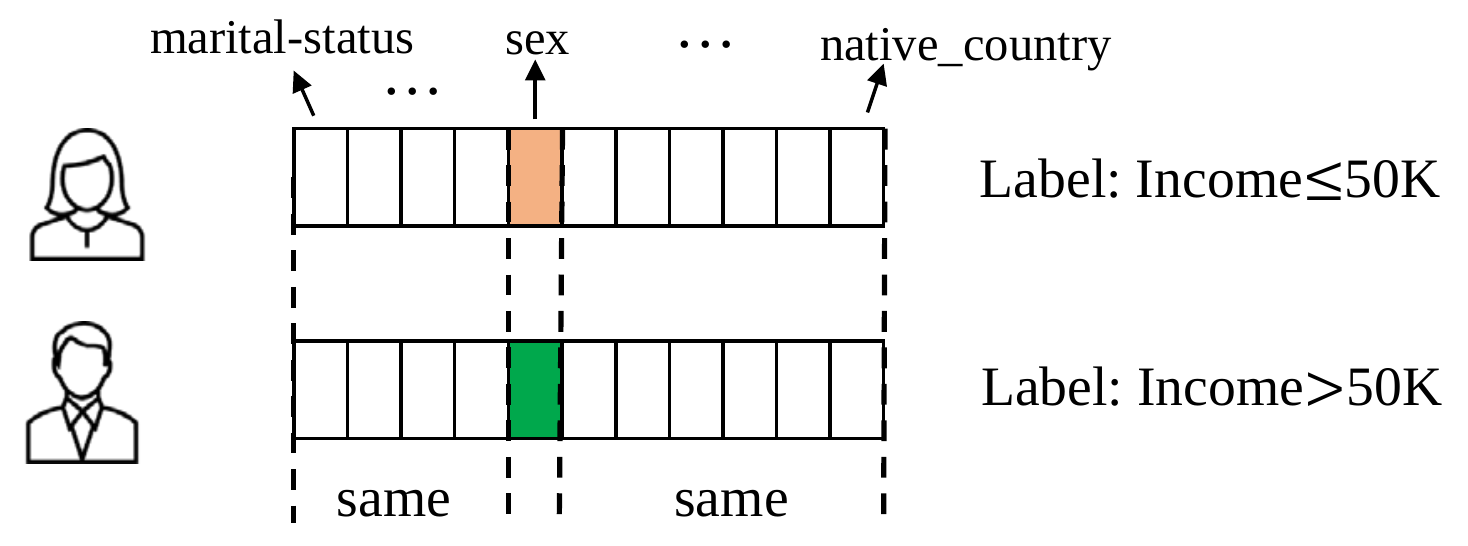}
	\caption{An example pair of discriminatory samples}
	\label{Fig-Sec2-Definition}
\end{figure}

In the rest of this paper: Section \ref{sec_Background} introduces the background and the problem definition. Section \ref{sec_Methodology} illustrates \mytool{}  and Section \ref{sec_evaluation} reports experimental results. Section \ref{sec_threats} and \ref{sec_relatedWork} discuss the threats to validity and related work. Section \ref{sec_conclusion} draws conclusions.

\section{Background and Problem Definition}
\label{sec_Background}


\subsection{Individual Discrimination}


There exists individual discrimination in a classifier if some protected attributes explicitly play important roles in the decision-making process~\cite{mitchell2018prediction, corbett2018measure}. The protected attributes are predefined in specific domains such as \textit{gender}, \textit{country}, and \textit{age}. 

Formally, given a classifier $f=<X,A>$, $X$ denotes the input domain, and $A = {A_1} \times {A_2} \times  \cdots  \times {A_n}$ represents the feature space of $X$. Each input sample $x \in X$ is represented as a \textit{n} dimensional feature vector, i.e., $x = \left\{ {{a_1},{a_2}, \ldots, {a_n}} \right\}$. Considering a set of protected attributes $P$, a classifier $f$ presents individual discrimination if there exists a pair of samples, $x$ and $x'$, satisfying the following Eq. (\ref{eq_proFeature})-(\ref{eq_labelRE}), where ${x=\{a_1, a_2, ..., a_n\}}$ and ${x^{'}} = \{ {{a_1^{'}},{a_2^{'}}, \ldots, {a_n^{'}}} \}$. In this case, $x$ and $x^{'}$ are  a pair of discriminatory samples in $f$. Here we use the symbols $\rhd$ and $\unrhd$ to denote that one feature is closely related and unrelated to a specific protected attribute, respectively.
\begin{equation}
	\small
	\exists a_i\in x, a_i\rhd p\in P, a_i\neq a'_i 
	\label{eq_proFeature}
\end{equation}
\begin{equation}
	\small
	\forall a_j\in x, a_j\unrhd p\in P, a_j=a'_j
	\label{eq_unproFeature}
\end{equation}
\begin{equation}
	\small
	f(x) \ne f({x^{'}})
	\label{eq_labelRE}
\end{equation}

Fig. \ref{Fig-Sec2-Definition} presents an example pair of discriminatory samples that are classified into different classes. The only difference between the two samples is their values of sex feature, which is related to the protected attribute \textit{gender}, indicating the unfairness of the classifier with regard to \textit{gender}. Note that the approach of ignoring certain protected attributes to mitigate model bias is not effective due to the presence of redundant encoding in the training dataset~\cite{pedreshi2008discrimination}.




\subsection{Locally Interpretable Method}

The drawbacks of invisible black-box models raise a series of security issues; many studies have proposed interpretable methods to measure the credibility of the prediction process. Among them, the locally interpretable methods~\cite{guo2018lemna,ribeiro2018anchors,ribeiro2016should,lundberg2017unified,guidotti2018local} are the most prominent. They interpret the predictions of black-box classifier models by providing potentially important features that dominate the decision results. By approximating the decision boundary of the classifier model, the generated interpreter learned with interpretable methods can explain the prediction specifically for a single sample. Concretely, given a test sample $x$, and its classification result $y=f(x)$, an interpreter $g$ will output explanation results $e$, a sorted set of features ranking based on their importance to this decision result.

\subsection{Genetic Algorithm}
The genetic algorithm (GA) is commonly used to search for approximate optimal solutions~\cite{mishra2019test,zhang2017nonlinear,sharma2016software}. GA first randomly initializes a population with potential solutions to the target problem. It then approximates the optimal solution iteratively through three biologically inspired operators, including selection, crossover, and mutation.

Specifically, GA encodes each solution in the initial population into a set of ``gene'' sequences and designs a fitness function to measure the qualities of current solutions. In the selection step, solutions with better fitness scores tend to be selected into the new population. In the crossover step, two parent solutions' gene fragments are exchanged to produce children solutions.  After that, a part of gene fragments of children solutions are mutated randomly, probably breaking out of the local search space and enhancing the diversity of solutions. GA repeats the whole  process until a \textit{desirable} solution is found or the number of iterations reaches the maximum as configured.

\subsection{Problem Definition}

For a given model, if the input is identified as an individual discriminatory sample, this model would be suffered from individual discrimination and may produce a biased decision. 
In this software fairness testing work, we assume that the dataset $X$ for training a model, the corresponding feature space $A$, and the protected attribute set $P$ are provided. The problem is that, given a black-box model $D$ trained using dataset $X$, can we effectively and efficiently detect individual discriminatory samples for $D$? By supplementing the discovered individual discriminatory samples to $X$, the model fairness is desired to have an improvement through retraining.  We have no prior knowledge (i.e., training algorithms and parameters) about the black-box model, making existing white-box based methods that require model inner information ineffective.

\section{Methodology}
\label{sec_Methodology}

\begin{figure}[!t]
	\centering
	\includegraphics[scale=0.32]{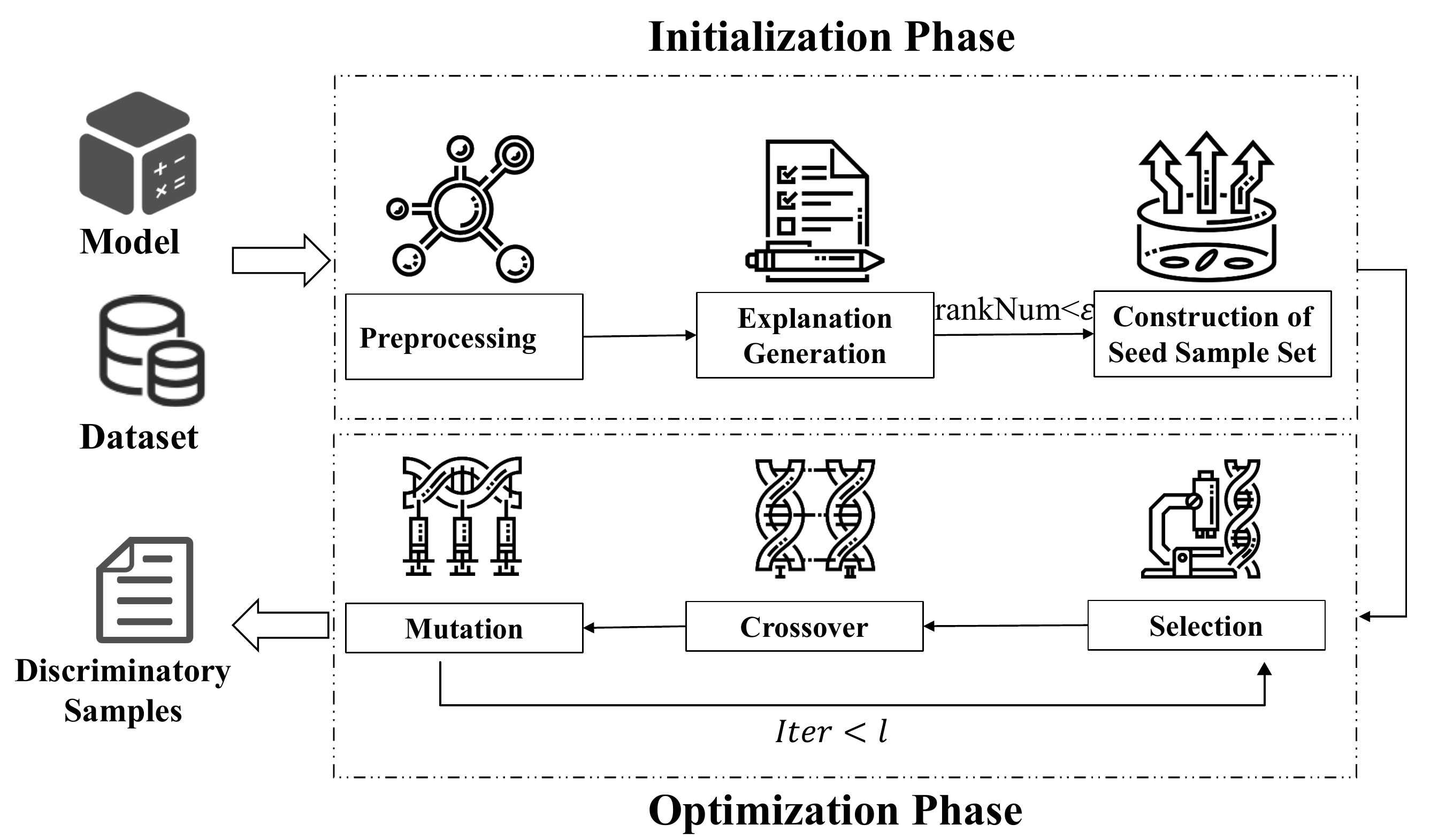}
	\caption{The overview of \mytool{}.}
	\label{Fig-Sec3-Overview}
\end{figure}
In this paper, we propose the \mytool{}, a lightweight model-agnostic software fairness testing method. Fig. \ref{Fig-Sec3-Overview} illustrates the overall framework of \mytool{}. The inputs are a dataset and its trained model. In the \textit{initialization phase}, \mytool{} employs an interpretable method to search for seed samples in the entire feature domain. Seed samples are more likely to derive discriminatory samples by slightly modifying feature values than other samples. Using seeds as inputs, the \textit{optimization phase} employs GA to efficiently generate a large number of discriminatory samples through the selection, crossover, and mutation operators.

\begin{figure}[!t]
	\centering
	\includegraphics[scale=0.4]{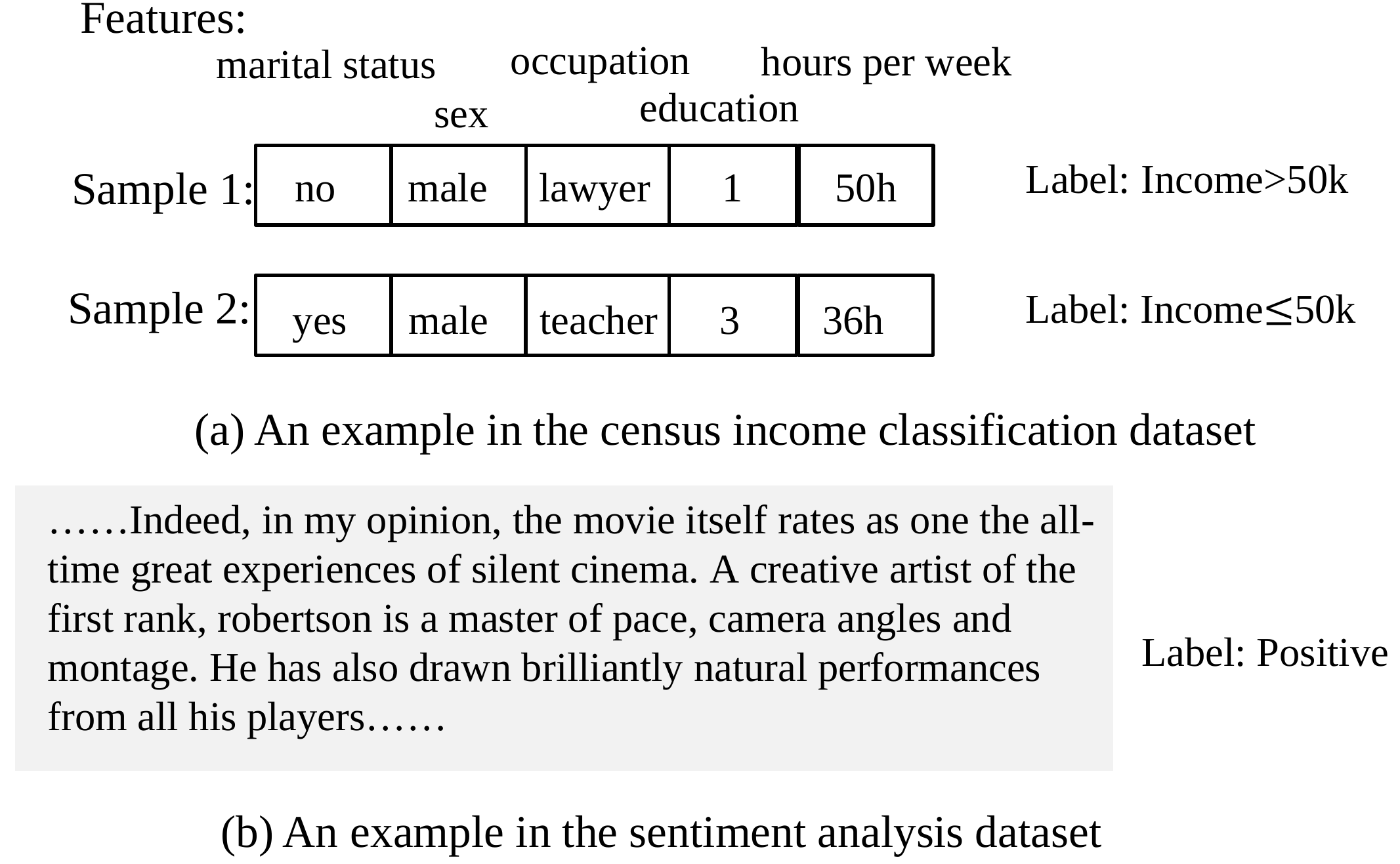}
	\caption{Two examples in tabular dataset and text dataset scenarios.}
	\label{Fig-Sec3-Example}
\end{figure}

To illustrate the generalizability of \mytool{}, we present two diverse scenarios that \mytool{} is able to handle: the input dataset is a tabular form or a set of text paragraphs. Fig. \ref{Fig-Sec3-Example} shows the two corresponding examples. The first example is selected from the census income classification dataset~\cite{Cencus}, as shown in Fig. \ref{Fig-Sec3-Example}(a). Each sample includes features of \textit{marital status}, \textit{sex}, \textit{occupation}, \textit{education}, \textit{hours per week}, etc. The labels of sample 1 and sample 2 are \textit{Income$>$50k} and \textit{Income$\le$50k}, respectively. Another example is selected from the sentiment analysis dataset for movie comments (i.e., IMDB dataset~\cite{IMDB}). As shown in  Fig. \ref{Fig-Sec3-Example}(b), each sample in this dataset is a text paragraph. 
Although the samples in the above two different scenarios are heterogeneous, they both can be consistently represented as feature vectors, $x=\{a_1,a_2,\dots,a_n\}$. For tabular samples,  $a_i$ denotes the  $i$-th feature value; for text samples,  $a_i$ denotes the $i$-th word. For convenience, we will uniformly use $a_i$ to denote the $i$-th word of sample $x$.

\subsection{Initialization Phase}
\label{subsec_globalSearch}

The initialization phase first identifies whether a sample contains sensitive words that are related to the given protected attributes $P$. Then it learns an interpreter by employing the interpretable methods, outputting the words that dominate  decision results. The samples whose sensitive word ranks high in these explanation results are selected as the seed samples. Concretely, this phase contains three steps, i.e., prepossessing, explanation generation, and construction of the seed sample set.

\subsubsection{Preprocessing} 
\label{subsubsec_preprocessing}
This step will identify the sensitive words that are related to the protected attributes. Considering that the protected attribute is $\{gender\}$, we can easily identify that the word ``sex'' in Fig. \ref{Fig-Sec3-Example}(a) is definitely close to the attribute since  they present a similar semantic meaning.
However, in some cases like Fig. \ref{Fig-Sec3-Example}(b), it is difficult to literally decide the relationships between the words and protected attributes due to the diversity of the language. In this example, the word ``master'' might be related to the gender information implicitly.

To solve this problem, we manually construct a knowledge graph containing the relations between sensitive words and protected attributes. At first, we obtain the protected attributes following an existing work~\cite{corbett2018measure,foulds2020intersectional, Protected}. Then, we construct a set of general relations, such as ``IsA'', ``RelatedTo'', ``DistinctFrom'', according to the ConceptNet~\cite{ConceptNet}, a widely used, open, and multilingual knowledge graph. After that, we expand the graph by adding similar words with a word embedding tool such as \texttt{Glove}~\cite{pennington2014glove}. For a new word, if there exists a similar word in the graph of which the similarity surpasses the configured threshold (i.e., 0.7), it will be added to the graph.  Finally, once a word has an accessible path to the protected attribute in the graph, it is regarded as a sensitive word that is related to the specific protected attribute. Note that the knowledge graph only needs to be constructed once.

\begin{figure}[!t]
	\centering
	\includegraphics[scale=0.37]{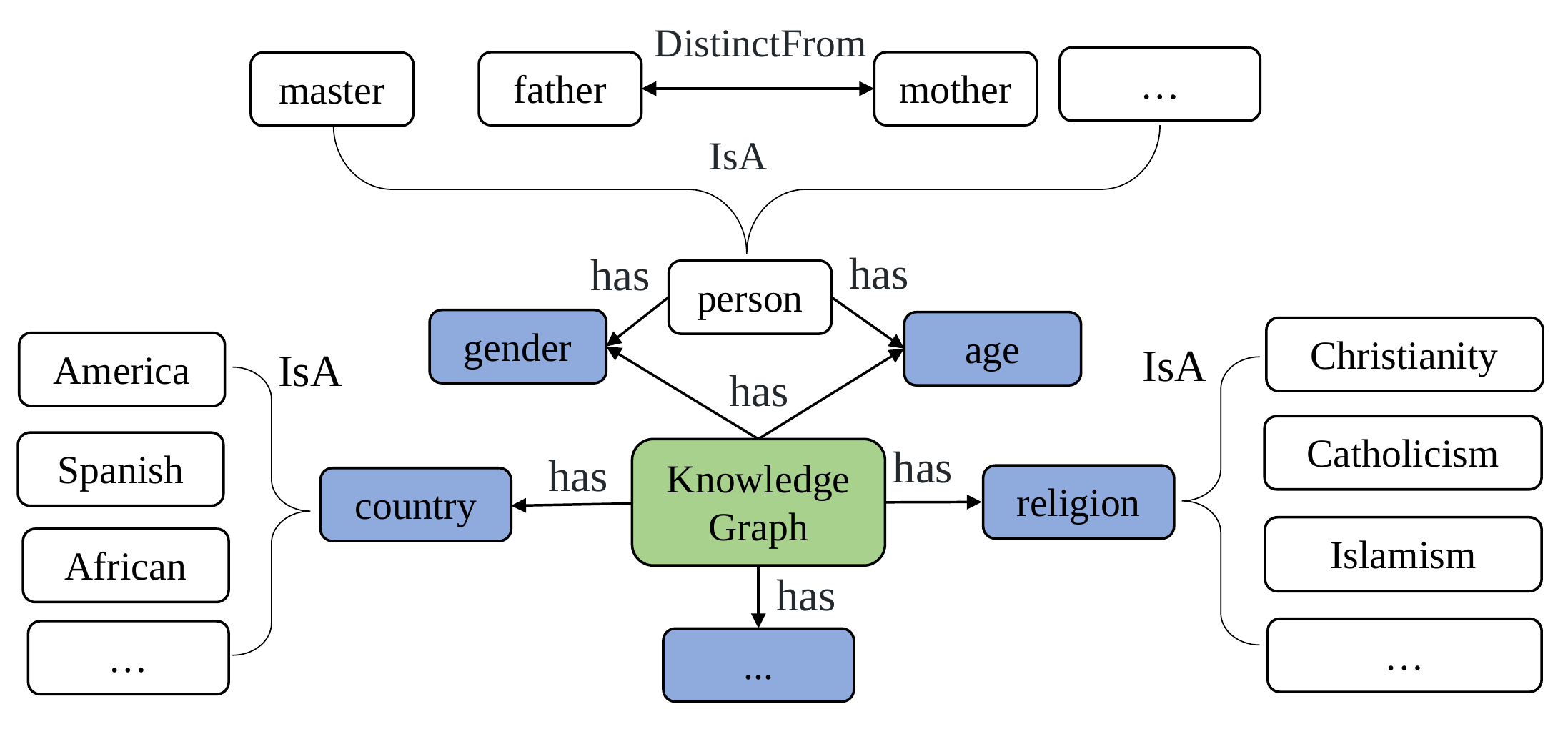}
	\caption{A partial prior knowledge graph.}
	\label{Fig-Sec3-Graph}
\end{figure}

Fig. \ref{Fig-Sec3-Graph} depicts a sub-graph of the knowledge graph we have constructed. The sub-graph contains a subset of protected attributes, such as \textit{gender}, \textit{age}, \textit{country}, and \textit{religion}. An edge of this graph  indicates a relation from a subject to an object. For instance, America ``IsA" a country. Based on this graph, for the example shown in Fig. \ref{Fig-Sec3-Example}(b), we can identify that ``master'' is a person and it has the gender attribute. Thus, it is identified as a sensitive word related to gender, that is, $master\rhd gender$. 

\subsubsection{Explanation Generation} 
Our assumption of using interpretable methods is that non-sensitive words should have a synergistic influence, dominating the classification of samples without discrimination. On the contrary, for discrimination samples, their sensitive words would rank relatively high in explanation results. As a result, a slight perturbation of sensitive words on discrimination samples might dramatically change the predicted label. Furthermore, we rely on the model-agnostic interpretable methods since they are good at training an interpreter for an opaque model, thus benefiting the generalization of \mytool{} to different models such as CNN, MLP, and SVM.
 
\begin{figure}[!t]
	\centering
	\includegraphics[scale=0.4]{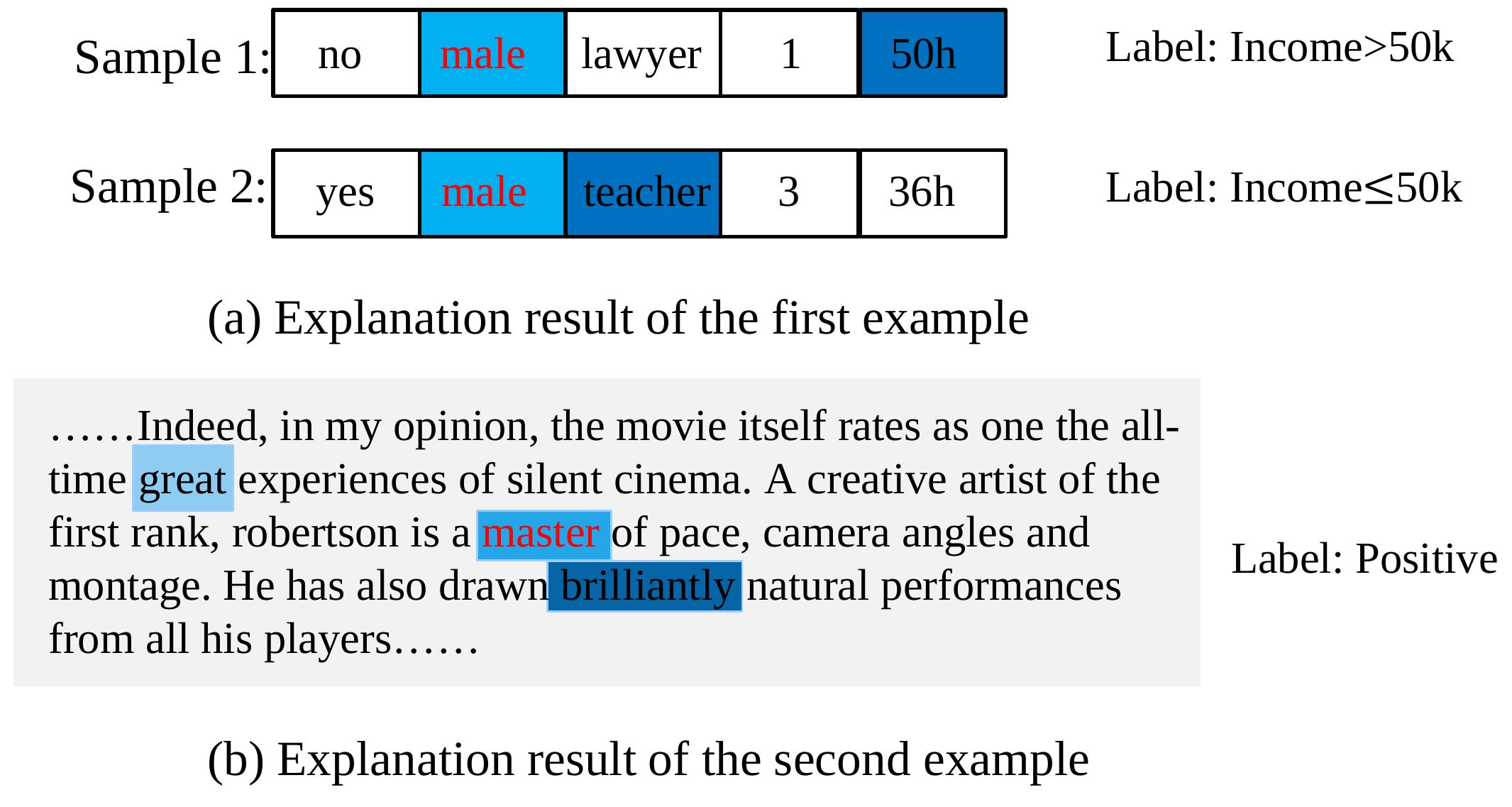}
	\caption{Explanation results of the two examples.}
	\label{Fig-Sec3-Explanation}
\end{figure}

Fig. \ref{Fig-Sec3-Explanation} presents the explanation results for the two examples in Fig. \ref{Fig-Sec3-Example}. The words ``male'' and ``master'' are marked with red color since they are identified as sensitive words during the prepossessing step.  The darker blue shading words play more essential roles in the classification process than the lighter shading words.


 \begin{algorithm2e}[t]
	\footnotesize
	\setcounter{AlgoLine}{0}
	\caption{Initialization phase}
	\label{GlobalSearch}
	\DontPrintSemicolon
	\SetCommentSty{mycommfont}
	\KwIn{$X$, $P$, Classifier $f$, Interpreter $g$, $seed\_num$, $\epsilon$, }
	\KwOut{Seed Sample Set $SeedSet$}
	$SeedSet=\emptyset$\;
	\ForEach{sample $x$ in $X$}{
	   \If{$|SeedSet|\le seed\_num$}{
	       $label=f(x)$\;
	       $e=g(x,f,label)$\; 
	       \ForEach{$a_i$ in $x$}{
	            \If{$a_i\rhd p, p\in P$}{
	                 $r=rankNum(a_i,e)$\;
	                 \If{$r\le \epsilon$}{
	                     $SeedSet = SeedSet\cup\{x\}$
	                 }
	            }
	       }
	   }
	}
	\KwRet{$SeedSet$}
\end{algorithm2e}

\subsubsection{Construction of Seed Sample Set}

This step constructs a set of seed samples, from which more discriminatory samples will be derived by the genetic algorithm. We first rank the explanation result $e$ based on the importance score assigned by the interpretable method. Then we design a function $rankNum(a_i,e)$ to obtain the rank number of a word $a_i$ in $e$. By iteratively obtaining the rank number of each sensitive word, we select the highest number to check whether it is higher than a threshold value $\epsilon$. If higher, the sample will be added to the seed sample set.

$\epsilon$ controls the balance between the number and quality of the selected seed samples. In Fig. \ref{Fig-Sec3-Explanation}, given that $\epsilon=2$, all samples are selected as seed samples since the rank numbers of sensitive words contained in these samples are 2. However, if $\epsilon=1$, none of them would be selected. It can be seen that a bigger $\epsilon$ would result in more seed samples; however, the rank number of sensitive words might be lower, making the sensitive words less important. Therefore, the quality of these samples would be poor. On the contrary, a smaller $\epsilon$ can result in high-quality seed samples with a smaller number, which might affect the efficiency of the generation of final discriminatory samples.

Algorithm \ref{GlobalSearch} formalizes the initialization implementations. The inputs include the dataset $X$, protected attribute set $P$, an interpreter $g$ constructed by leveraging the existing interpretable method, and the target number of seed samples ($seed\_num$). The output is a set of seed samples, denoted as $SeedSet$. Each sample is first fed to the classifier and interpreter to generate its explanation results (lines 2-5). The sample, which contains any sensitive word with a smaller ranking number than  $\epsilon$, will be added to $SeedSet$ (lines 6-10).

\subsection{Optimization Phase}
\label{subsec_localGeneration}

\begin{algorithm2e}[t]
	\footnotesize
	\setcounter{AlgoLine}{0}
	\caption{Optimization phase}
	\label{LocalGeneration}
	\DontPrintSemicolon
	\SetCommentSty{mycommfont}
	\KwIn{$SeedSet$, $l$}
	\KwOut{Discriminatory Sample Set $DisSet$} 
	$CurPopulation=InitPopulation(SeedSet)$\;
	$num=0$\;
	\While{$num\le l$}{
		$NewPopulation=Selection(CurPopulation)$\;
		$CurPopulation=Crossover(CurPopulation)$\;
		$CurPopulation=Mutator(CurPopulation)$\;
		\ForEach{sample $x$ in $CurPopulation$}{
			\If{$True==DiscrniminatoryCheck(x)$}{
				$DisSet=DisSet\cup \{x\}$}}
	}
	\KwRet{$DisSet$}
\end{algorithm2e}

The optimization phase leverages the biologically inspired GA to generate discriminatory samples, taking advantage of its high convergence speed and strong local searching ability. 
Here we use Algorithm \ref{LocalGeneration} to help understand the order of the three operators. At first, we will construct the initial population based on the $SeedSet$ input (line 1). Then, the complete process of the selection, crossover, and mutation (lines 4-6) will be repeated $l$ times, as set by users. After finishing each iteration, we leverage $DiscrniminatoryCheck()$ to check  whether a candidate sample $x$ is a \textit{true} discriminatory sample. If the output set $DisSet$ does not contain the sample, $x$ will be added to the set (lines 7-9). 
The details of the above steps are introduced below.  

\subsubsection{Construction of Initial Population}

This step constructs the initial population based on seed samples. However, it is different to construct the initial populations for the tabular dataset and the text dataset. 

For seed samples from a tabular dataset, all of them will be added to the initial population since they have the same number of features.  

For seed samples from a text dataset, it is unfeasible to initialize the population by simply aggregating all seed samples since the lengths of paragraphs in different text samples are diverse. 
A simple replacement of a word $a_i$ in one sample with that word $a'_i$ in another sample (in the next crossover step) would lead to a new sentence with syntax errors or incorrect semantic. In this work, we construct an initial population correspondingly from each seed sample. 
Specifically, for a given seed sample, we first randomly select $k$ non-sensitive words from it. Correspondingly, $k$ new words that have similar meanings with those $k$ selected words can be obtained through the similarity computation by the word embedding tool \texttt{Glove}. 
After that, we replace the originally selected words with these new words, producing $k$ new samples derived from the seed sample. The set of these $k+1$ samples will form an initial population. Our work sets $k=20$ according to experimental experience. 


\subsubsection{Selection}
\label{subsubsec_selection}

This step selects high-quality samples to generate the next population through optimizing a fitness function.  We design a fitness function $fit(x)$ for a sample $x$. $fit(x)$ quantifies the extent of a change from the prediction probability of $x'$ to that of $x''$. $x'$ and $x''$ are derived from $x$. We assume that $x$ with a higher fitness score is more likely to be mutated into a discriminatory sample. $fit(x)$ can be obtained with Eq. (\ref{eq-fitness}):
\begin{equation}
\small
fit(x)=|Prob(x',l)-Prob(x'',l)|
\label{eq-fitness}
\end{equation}

We now introduce how to derive $x'$ and $x''$ from $x$. Given that $x$ contains a sensitive word $a_i\rhd p$, we construct the $<\widetilde{a_i}, \neg\widetilde{a_i}>$ that is a pair of sensitive words but with opposite semantic meanings.  
We then substitute $\widetilde{a_i}$ and $\neg\widetilde{a_i}$ for the original $a_i$, producing two new samples, $x'$ and $x''$, respectively. $Prob(x',l)$ and $Prob(x'',l)$ correspond to the prediction probability of $x'$ and $x''$ for the specific label $l$. 

For example, given $a_i$=``actor'' and $p$=\textit{gender}, $\widetilde{a_i}$ is the same as $a_i$, and $\neg\widetilde{a_i}$ would be ``actress''. Another example is that, when $a_i$= ``master", $p$=\textit{gender}, and $a_i$ has a relationship with $p$, we would set $\widetilde{a_i}=$``male master" and $\neg\widetilde{a_i}$=``female master" based on the prior knowledge graph we have constructed in Section \ref{subsubsec_preprocessing}. 

Algorithm \ref{Selection} shows the selection in terms of $fit(x)$. The input is current population $CurPopulation$ and the output is a new population $NewPopulation$ after selection.
$sumFit$ denotes the sum of all fitness scores of the samples in  $CurPopulation$. $\delta(x)$ denotes the probability value of $x$ that might be selected into the new population.
$\Delta(X)$ is a set of all $\delta(x)$ of samples in $CurPopulation$. Finally, selecting based on the probability distribution indicated from $\Delta(X)$, we can construct a new population that has the same sample size as $Curpopulation$. 

 \begin{algorithm2e}[t]
	\footnotesize
	\setcounter{AlgoLine}{0}
	\caption{Selection of new population}
	\label{Selection}
	\DontPrintSemicolon
	\SetCommentSty{mycommfont}
	\KwIn{$CurPopulation$}
	\KwOut{$NewPopulation$}
	$NewPopulation=\emptyset$\;
	$sumFit=\sum_{x\in CurPopulation}{fit(x)}$\;
	$\Delta(X)=\emptyset$\;
	\ForEach{sample $x$ in $CurPopulation$}{
	    $\delta(x)=fit(x)/sumFit$\;
	    $\Delta(X)=\Delta(X)\cup \delta(x)$\;
	}
	\While{$|NewPopulation|<|CurPopulation|$}{
	    $z=select(CurPopulation,\Delta(X))$\;
	    $NewPopulation=NewPopulation\cup z$     
	}
	\KwRet{$NewPopulation$}
\end{algorithm2e}

 \begin{algorithm2e}[t]
	\footnotesize
	\setcounter{AlgoLine}{0}
	\caption{Crossover and Mutation}
	\label{CrossoverAndMutation}
	\DontPrintSemicolon
	\SetCommentSty{mycommfont}
	\KwIn{$CurPopulation$, $cr$, $mr$}
	\KwOut{$CurPopulation$}
	\ForEach{sample $x$ in $CurPopulation$}{
	    $x'=randomSelect(CurPopulation)$\;
	    $<a_i,\dots, a_j>=randomSelect(x, cr)$\;
	    $frag(x,i,j)=<a_i,\dots, a_j>$\;
	    $x=x.replace(frag(x,i,j),frag(x',i,j))$\;
	    $x'=x'.replace(frag(x',i,j),frag(x,i,j))$\;
	}
	\ForEach{sample $x$ in $CurPopulation$}{
	    \ForEach{$a_i$ in $x$}{
	        \If{$a_i\unrhd p$ }{//$\unrhd$ denotes that $a_i$ is not related to $p$\;
	           $a'_i=calSimilarWord(a_i,mr)$\;
	           $x=x.replace(a_i, a'_i)$\;
	        }
	    }
	}
	\KwRet{$CurPopulation$}
\end{algorithm2e}

\subsubsection{Crossover and Mutation}
We conduct the crossover and mutation operators on the selected population to generate more varieties, expanding the search space of discriminatory samples. Algorithm \ref{CrossoverAndMutation} illustrates the crossover (lines 1-6) and the mutation (lines 7-12). The two parameters, $cr$ and $mr$, denote the crossover rate and mutation rate.   

We randomly choose pairs of samples as the crossover parents from the current population with a probability of $cr$. For a pair of samples, $x$ and $x'$, fragments $<a_i,\dots, a_j>$ and $<a'_i,\dots, a'_j>$ are extracted from $x$ and $x'$ randomly. The two samples will exchange their fragments with each other.

During the mutation, each word of a sample in the current population is replaced with its similar word with a probability of $mr$. Note that the mutation excludes the sensitive words. Concretely, 1) categorical word such as ``no'' and ``teacher'' will be replaced with similar words in the same category such as ``yes'' and ``lawyer''; 2) text word  such as ``great'' and ``brilliantly'' will be replaced with similar words using the tool \texttt{Glove}; 3) numeric word such as ``1'' and ``36h'' can be replaced with numeric data like ``2'' and ``40h''. 

\begin{algorithm2e}[t]
	\footnotesize
	\setcounter{AlgoLine}{0}
	\caption{Checking of discriminator sample}
	\label{DiscriminatorChecking}
	\DontPrintSemicolon
	\SetCommentSty{mycommfont}
	\KwIn{$x$, Classifier $f$, $P$}
	\KwOut{$Dis$}
	$Dis=false$\;
	\If{$\exists a_i\in x, a_i\rhd p, p\in P$}{
		$<\widetilde{a_i}, \neg\widetilde{a_i}>=getPair(a_i)$\;
		$x'=x.replace(a_i,\widetilde{a_i})$\;
		$x''=x.replace(a_i, \neg\widetilde{a_i})$\;
		\If{$f(x')\neq f(x'')$}{
			$Dis=true$\;
		}
	}
	\KwRet{$Dis$}
\end{algorithm2e}

\begin{figure}[!t]
	\centering
	\includegraphics[scale=0.3]{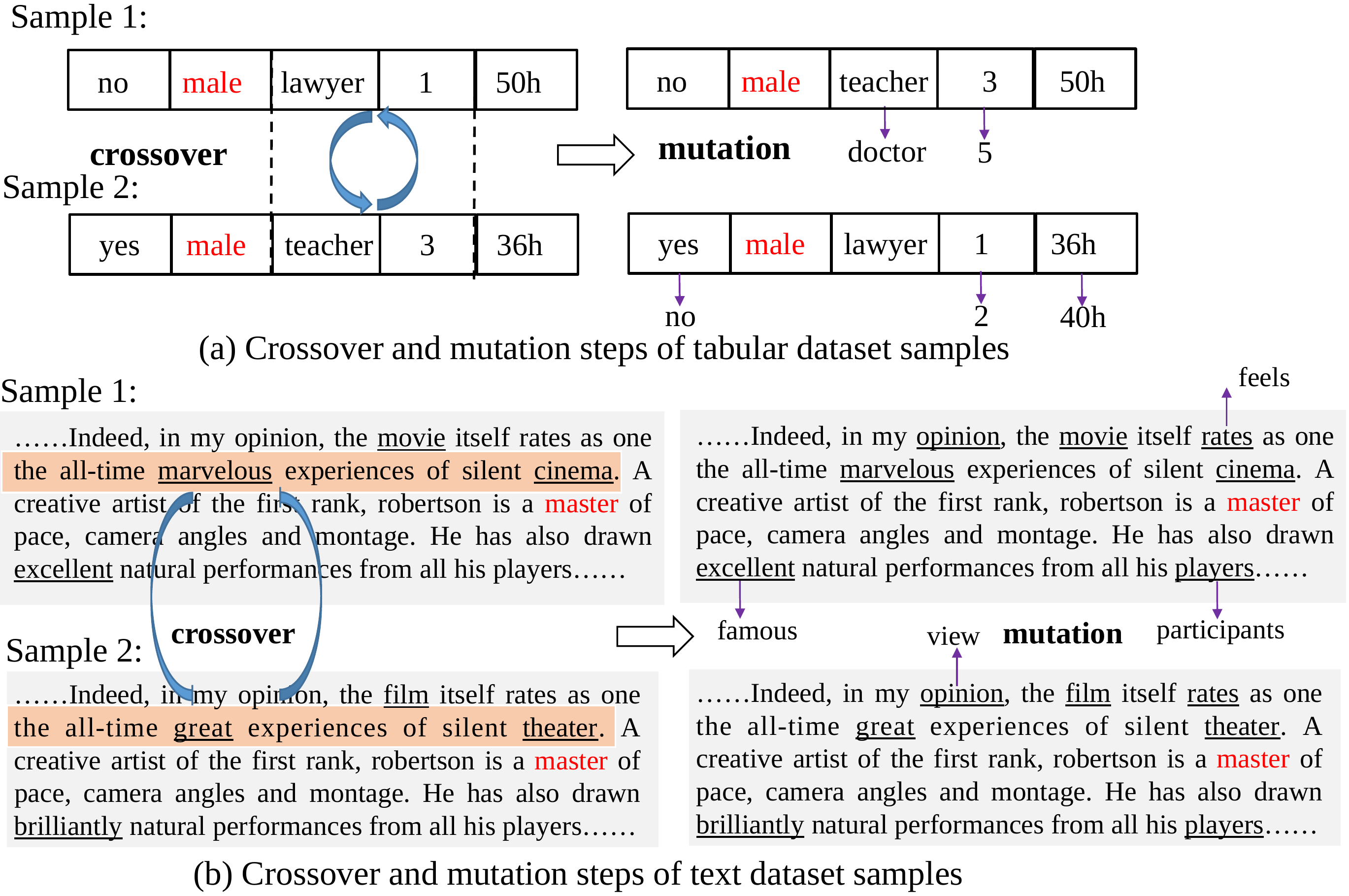}
	\caption{Examples for the crossover and mutation steps.}
	\label{Fig-Sec3-CrossoverMutation}
\end{figure}

Fig. \ref{Fig-Sec3-CrossoverMutation}(a) shows the crossover and mutation for the tabular samples and Fig. \ref{Fig-Sec3-CrossoverMutation}(b) for the text samples. In Fig. \ref{Fig-Sec3-CrossoverMutation}(a),  the crossover fragments are $<$``lawyer'',``1''$>$ in sample 1 and $<$``teacher'',``3''$>$ in sample 2. Then the words of ``teacher'', ``3", ``yes", ``1", and `"36h" are  mutated with  ``doctor'', ``5", ``no'', ``2", and ``40h".  In \ref{Fig-Sec3-Explanation}(b),  the different words between the two samples are underlined. The two sub-sentences in the orange shading are exchanged with each other. After that, some words like ``excellent"  will be mutated with  their similar words like ``famous".


\subsubsection{Checking of Discriminatory Sample} After each iteration of the selection, crossover, and mutation, we check whether a given sample is a discriminatory one by identifying their label difference using Algorithm \ref{DiscriminatorChecking}. Specifically, once a sample contains a sensitive word $a_i$, we leverage the function $getPair$ to return $<\widetilde{a_i}, \neg\widetilde{a_i}>$. Then, we replace the word $a_i$ with the new words and obtain two new samples. Once the two new samples' predicted labels are different, the given sample will be checked as a discriminator one.

\section{Evaluation}
\label{sec_evaluation}

In this section, we evaluate the performance of \mytool{} on both the tabular datasets and text datasets. We first introduce our experimental setup (Section \ref{subsec_setup}). Then we conduct the evaluation by answering the  three research questions. 

\emph{\textbf{RQ1}: What is the performance of \mytool{} in finding discriminatory samples on tabular datasets? (Section \ref{subsec_tabular})} 

\emph{\textbf{RQ2}: What is the performance of \mytool{} in finding discriminatory samples on text datasets? (Section \ref{subsec_text})} 

\emph{\textbf{RQ3}: To what extent do the discriminatory samples generated by \mytool{} improve the model fairness through retraining? (Section \ref{subsec_retraining})}

\subsection{Experimental Setup}
 \label{subsec_setup}

\subsubsection{Datasets}
\label{subsubsec_dataset}
We conduct the evaluation on five popular public benchmark datasets, including three tabular datasets and two text datasets. Their descriptions are listed below:
\begin{itemize}
	\item {\textbf{Census Income Dataset~\cite{Cencus}} labels whether the income of an adult is over \$50K or not. There are  32,561 samples and 13 features, involving  three protected attributes, i.e., \textit{gender}, \textit{age}, and \textit{race}. }
	
	\item{\textbf{German Credit Dataset~\cite{Credit}} is used to assess the credit level of applicants (i.e., good or bad) based on their personal conditions. There are 600 samples depicted by 20 diverse features. The protected attributes are \textit{gender} and \textit{age}.}
	
	\item{\textbf{Bank Marketing Dataset~\cite{Bank}} is related to the direct marketing campaign of a Portuguese banking institution. The objective is to predict whether the client would subscribe to a term deposit based on their information. There are 16 features and 45,211 samples. The protected attribute is \textit{age}.} 
	
	\item{\textbf{IMDB Dataset~\cite{IMDB}} is a large movie review dataset for binary sentiment classification. The original dataset contains 50,000 samples, where each sample contains 210 words on average. The protected attributes are \textit{gender} and \textit{country}}.
	
	\item{\textbf{SST Dataset~\cite{SST}} is also used to classify whether a movie comment is positive or negative. The original dataset contains 9,613 samples, where each one has about 19 words on average. The protected attributes are the same as those of the IMDB dataset. }
\end{itemize}

For convenience, we use Cencus, Credit, and Bank to denote the abbreviations of the three tabular datasets. 

 
 \subsubsection{Baseline Approaches}
 \label{subsubsec_baseline}
 On the tabular datasets, we select three state-of-the-art approaches, i.e., AEQUITAS, SG, and ADF, as the baseline approaches. Their descriptions are listed below:
 \begin{itemize}
 	\item{\textbf{AEQUITAS~\cite{udeshi2018automated}}:} It is a two-phase method to search for individual discriminatory samples, where the search is random in its global phase and is directed in its local phase. AEQUITAS proposes three schemes, i.e., random scheme, semi-directed scheme, and fully-directed scheme, to guide the local stage.  A fully-directed scheme performs best, which is adopted in our comparative experiment.
 	\item{\textbf{SG~\cite{aggarwal2019black}}:} SG is based on program symbolic execution and interpretable method. Instead of directly obtaining explanation results for a single sample, SG uses LIME to generate local perturbation samples to build a decision tree and generates test inputs by solving path constraints. 
 	\item{\textbf{ADF~\cite{zhang2020white}}:} It searches for seed discriminatory sample pairs along with the gradient direction and generates more discriminatory samples locally guided by gradients in an opposite way. 
 \end{itemize}
 
 In our experiments, we implement AEQUITAS and ADF based on their public repositories published on Github~\cite{AequitasCode, ADFCode}. We implement SG based on the refactored version available in the ADF repository. The parameter settings in experiments are the same as the optimal settings given in their papers.
 
  On the text datasets, we use MT-NLP as the baseline approach.  
   \begin{itemize}
   	   \item{\textbf{MT-NLP~\cite{ma2020metamorphic}} produces test inputs according to Metamorphic Testing (MT). MT-NLP uses a knowledge graph to locate sensitive words in a sentence. Afterward, it employs two mutation schemes to generate perturbed sentences, followed by a discriminatory checking on all these perturbed sentences. }
   	\end{itemize}

Given that our proposed approach and the baselines contain random factors when selecting seed samples, we conduct the experiment 30 times and use the average values on tabular datasets.  Note that for the MT-NLP, since its source code is not available, we conduct the comparison using its paper results.  
 
\subsubsection{Trained Models and Interpreters}
\label{subsubsec_models}

\begin{table}[!t]
\newcommand{\tabincell}[2]{\begin{tabular}{@{}#1@{}}#2\end{tabular}}
\caption{Training parameters of different models.}
\label{Table-ModelParameter}
\centering
\scalebox{0.95}{
\begin{tabular}{l|l|l} 
\toprule[1.5pt]
Dataset & Model & Training Parameters   \\ 
\hline
Cencus      & SVM   & rbf kernel                            \\ 
\cline{2-3}
Credit       & RF    & 100 trees                       \\ 
\cline{2-3}
Bank       & MLP   & \tabincell{l}{five-layer fully-connected NN\\ (128 neurons each layer)}                          \\
\hline
IMDB      & CNN   & \tabincell{l}{one convolution layer,\\ one pooling layer,\\  one fully-connected NN layer}                   \\ 
\cline{2-3}
SST      & LR    &   l2 penalty                            \\
\bottomrule[1.5pt]
\end{tabular}
}
\end{table}

To evaluate the generalizability of \mytool{}, we conduct \mytool{} on various models, which are also used in the baseline approaches. As shown in Table \ref{Table-ModelParameter}, we train the SVM, Random Forest(RF), and MLP models on tabular datasets, and train the CNN and Logistic Regression(LR) models on text datasets. Here the traditional machine learning models, i.e., RF, SVM, and LR models, are trained via scikit-learn~\cite{SKLearn} library, and the deep learning models such as MLP and CNN are trained via TensorFlow~\cite{TensorFlow}. Table \ref{Table-ModelParameter} lists key parameters for training these models. The other parameters absent in this table use the default values configured by scikit-learn and TensorFlow.

Based on the trained models, we leverage existing locally interpretable methods, LIME~\cite{ribeiro2016should} and SHAP~\cite{lundberg2017unified}, to generate corresponding interpreters.  Recent empirical studies~\cite{fan2020can, warnecke2020evaluating} on different interpretable methods present that LIME performs best on tabular datasets and SHAP outperforms the others on text datasets. Consistently, our experiments use LIME for tabular datasets and SHAP for text datasets by default.

\subsubsection{Measurement Metrics}
\label{subsubsec_metrics}

We use DSS and SUR metrics to measure the efficiency and effectiveness of discriminatory sample detection by different methods. We assume that a discriminatory sample detection method presents a better performance if it achieves a smaller value of DSS and a larger value of SUR. 

DSS denotes the average consumed time for generating a discriminatory sample,
\begin{equation}
\small
DSS=\frac{Time}{DSN}
\label{eq-DSS}
\end{equation}
where $Time$ and $DSN$ denote the consumed time and the number of detected discriminatory samples, respectively.
SUR denotes the success rate for generating discriminatory samples, 
\begin{equation}
\small
SUR=\frac{DSN}{TSN}
\label{eq-SUR}
\end{equation}
where TSN is the total number of generated testing samples. It is worth noting that when calculating the number of generated testing samples, we only consider the samples that are fed to the classifiers to perform discriminatory checking.

\subsubsection{Parameters Setting}
\label{subsubsec_parameter}

Recall that \mytool{} involves three important parameters, including $\epsilon$ for seed sample selection, $cr$ for the crossover, and $mr$ for the mutation. Table \ref{Table-ParameterSetting} lists these parameter values configured for each dataset. 

To configure an appropriate $\epsilon$ value for each dataset, we first randomly select 100 samples from each dataset and generate explanation results. Then, we rank these samples in increasing order according to their sensitive word ranking numbers. $\epsilon$ value is set as the ranking number of the $20^{th}$ sample, indicating that about 20\% of the samples are regarded as seed samples. We configure $cr$ and $mr$ according to our experimental experience. 

\begin{table}[!t]
	\newcommand{\tabincell}[2]{\begin{tabular}{@{}#1@{}}#2\end{tabular}}
	\caption{Parameter settings in different datasets.}
	\centering
	\scalebox{0.95}{
	\begin{tabular}{llll} \toprule[1.5pt]
		Dataset & $\epsilon$ & $cr$ & $mr$  \\  \hline
		Cencus       & 7  &  0.9 & 0.05   \\
		Credit       & 14  & 0.9  & 0.05   \\
		Bank       &  9 & 0.9  &  0.05  \\
		IMDB       &  20 & 0.5  &  0.05  \\
		SST       & 20  & 0.5  &  0.05   \\ \bottomrule[1.5pt]
	\end{tabular}
}
\label{Table-ParameterSetting}
\end{table}

Our experiments are conducted on a server with Ubuntu 18.04 operating system, Intel Xeon 2.50GHz CPU, NVIDIA RTX GPU, and 128GB system memory.


\subsection{Performance on Tabular Datasets}
\label{subsec_tabular}

 \begin{table*}[!t]
 
	\newcommand{\tabincell}[2]{\begin{tabular}{@{}#1@{}}#2\end{tabular}}
	\caption{The performance of detecting discriminatory samples for four approaches using MLP model.}
	\label{Table-DSS-MLP}
\centering
\scalebox{0.85}{
\begin{tabular}{ll|rrrr|rrrr|rrrr|rrrr} 
	 \toprule[1.5pt]
	\multicolumn{2}{c|}{\multirow{2}{*}{Dataset}} & \multicolumn{4}{c|}{ExpGA} & \multicolumn{4}{c|}{AEQUITAS} & \multicolumn{4}{c|}{SG} & \multicolumn{4}{c}{ADF}  \\
	\multicolumn{2}{c|}{}                &  TSN &   DSN     & DSS  &  SUR              &  TSN  &   DSN    & DSS &  SUR      & TSN   &  DSN   & DSS   &  SUR   &   TSN   &  DSN    & DSS   &  SUR             \\ 
	\hline
	\multirow{3}{*}{Census} & gender      &  \textbf{491,323}   &   \textbf{141,049} & \textbf{0.03}   &  \textbf{28.70\%}     &   85,629 &  1,797  & 2.00   &  2.10\%  & 7,070     &  871   & 4.13 & 12.31\%       &  116,119    &    24,974 & 0.14   &    21.51\%           \\
	& age                 &   \textbf{43,465}    &   \textbf{29,467}  & \textbf{0.12}   &   \textbf{67.79\%}       & 12,243   &   1,107  & 3.25 &   9.04\%     &   4,190    &   1,530 & 2.35  & 36.51\%              &   40,135      &   21,795 & 0.17  &   54.30\%           \\
	& race                &  \textbf{90,303}  &  \textbf{30,171}  & \textbf{0.12}  &    \textbf{33.41\%}      &  22,053  &    551    & 6.53  &  2.50\%              &   4,355   &   588   & 6.12  &  13.50\%            &    47,296  &  8,097 & 0.44  & 17.12\%             \\ 
	\hline
	\multirow{2}{*}{Credit} & gender              & \textbf{281,130}   &  \textbf{65,009}   & \textbf{0.06}  & \textbf{23.12\%}     & 38,096  &   1,596  & 2.26  & 4.19\%        &   7,569 &  1,321 & 2.73 &  17.45\%           &  62,242  &  8,052 & 0.45  &  12.93\%          \\
	& age                 &  \textbf{46,357}  &   \textbf{34,048}  & \textbf{0.11}  &  \textbf{73.44\%}     & 11,756   &  5,146 & 0.70  &  43.77\%                   & 7,699  &  4,659 & 0.77 &  60.5\%           & 21,649   &  5,957 & 0.60  &  27.51\%           \\ 
	\hline
	Bank           & age   & \textbf{46,802}  & \textbf{27,099}  & \textbf{0.13} &   \textbf{57.90\%}       & 14,808   &  2,915  & 1.23  &  19.68\%                 &  3,776 &  2,592 & 1.39 &   68.63\%          & 27,325  &   11,852  & 0.30  &   43.37\%         \\
	\bottomrule[1.5pt]
\end{tabular}
}
\end{table*}

 \begin{table*}[!t]
	
	\newcommand{\tabincell}[2]{\begin{tabular}{@{}#1@{}}#2\end{tabular}}
	\caption{The performance of detecting discriminatory samples for three approaches  using RF model.}
	\label{Table-DSS-RF}
\centering
\scalebox{0.87}{
\begin{tabular}{ll|rrrr|rrrr|rrrr} 
	\toprule[1.5pt]
	\multicolumn{2}{c|}{\multirow{2}{*}{Dataset}}         & \multicolumn{4}{c|}{ExpGA}                      & \multicolumn{4}{c|}{AEQUITAS}                  & \multicolumn{4}{c}{SG}            \\
	\multicolumn{2}{l|}{}                                 & TSN     & DSN     & DSS  & SUR                  & TSN    & DSN    & DSS   & SUR                  & TSN    & DSN    & DSS  & SUR      \\ 
	\hline
	\multirow{3}{*}{Census} & gender               & \textbf{274,521} & \textbf{156,259} & \textbf{0.02} & \textbf{56.92\%}              & 56,776 & 1,210  & 2.98  & 2.13\%               & 10,906 & 595    & 6.05 & 5.46\%   \\
	& age                  & \textbf{51,192}  & \textbf{46,035}  & \textbf{0.08} & \textbf{89.93\%}              & 21,330 & 13,472 & 0.27  & 63.16\%              & 17,400 & 12,441 & 0.29 & 71.50\%  \\
	& race                 & \textbf{78,349}  & \textbf{47,810}  & \textbf{0.08} & \textbf{61.02\%}              & 29,309 & 362    & 9.94  & 1.24\%               & 12,572 & 894    & 4.03 & 7.11\%   \\ 
	\hline
	\multirow{2}{*}{Credit} & gender               & \textbf{20,295}  & \textbf{4,247}   & \textbf{0.85} & 20.93\%              & 3,245  & 148    & 24,32 & 4.56\%               & 7,269  & 2,192  & 1.64 & \textbf{30.15\%}  \\
	& age                  & \textbf{10,183}  & \textbf{3,673}   & \textbf{0.98} & 36.07\%              & 869    & 153    & 23.53 & 17.60\%              & 6,349  & 2,654  & 1.36 & \textbf{41.80\%}  \\ 
	\hline
	Bank                 & age                  & \textbf{13,284}  & \textbf{8,885}   & \textbf{0.41} & \textbf{66.99\%}              & 3,952  & 941    & 3.83  & 23.81\%              & 3,871  & 2,530  & 1.42 & 65.36\%  \\
	 \bottomrule[1.5pt]
     
\end{tabular}
}
\end{table*}

 \begin{table*}[!t]
	
	\newcommand{\tabincell}[2]{\begin{tabular}{@{}#1@{}}#2\end{tabular}}
	\caption{The performance of detecting discriminatory samples for three approaches using SVM model.}
	\label{Table-DSS-SVM}
	\centering
	\scalebox{0.87}{
\centering
\begin{tabular}{ll|rrrr|rrrr|rrrr} 
	\toprule[1.5pt]
	\multicolumn{2}{c|}{\multirow{2}{*}{Dataset}} & \multicolumn{4}{c|}{ExpGA}         & \multicolumn{4}{c|}{AEQUITAS}     & \multicolumn{4}{c}{SG}           \\
	\multicolumn{2}{c|}{}                         & TSN     & DSN     & DSS  & SUR     & TSN     & DSN   & DSS   & SUR     & TSN   & DSN   & DSS   & SUR      \\ 
	\hline
	\multirow{3}{*}{Census} & gender       & \textbf{804,328} & \textbf{145,226} & \textbf{0.02} & \textbf{18.06\%} & 109,706 & 386   & 9.33  & 0.35\%  & 1,119     & 0     & -     & 0\%      \\
	& age          & \textbf{124,816} & \textbf{78,064}  & \textbf{0.05} & \textbf{62.54\%} & 81,275  & 2,489 & 1.45  & 3.06\%  & 2,998 & 624   & 5.77  & 20.81\%  \\
	& race         & \textbf{226,434} & \textbf{78,627}  & \textbf{0.05} & \textbf{34.72\%} & 62,954  & 277   & 13.0  & 0.44\%  & 1,141     & 0     & -     & 0\%      \\ 
	\hline
	\multirow{2}{*}{Credit} & gender       & \textbf{118,428} & \textbf{47,570}  & \textbf{0.08} & \textbf{40.17\%} & 152,432 & 3,683 & 0.98  & 2.42\%  & 9,204 & 1,912 & 1.88  & 20.77\%  \\
	& age          & \textbf{21,058}  & \textbf{9,377}   & \textbf{0.38} & \textbf{44.53\%} & 53,344  & 8,419 & 0.43  & 15.78\% & 7,952 & 1,918 & 1.88  & 24.12\%  \\ 
	\hline
	Bank                  & age          & \textbf{42,746}  & \textbf{29,574}  & \textbf{0.12} & \textbf{69.19\%} & 31,276  & 41    & 87.80 & 0.13\%  & 744   & 248   & 14.52 & 33.33\%  \\
	\bottomrule[1.5pt]
\end{tabular}
}
\end{table*}

To answer RQ1, we compare \mytool{} with AEQUITAS, SG, and ADF on three tabular datasets. For the Census and Bank datasets, we first randomly select 1,000 samples from each as the input. For the  Credit dataset that contains 600 samples, we input all the samples into the methods. Then, we insert a timer in each method to count the generated discriminatory samples every five minutes during a period of one hour. After that, we measure TSN, DSN, DSS, and SUR for each method. 

Table \ref{Table-DSS-MLP}, Table \ref{Table-DSS-RF}, and Table \ref{Table-DSS-SVM} present performance measurements of the four methods based on MLP, RF, and SVM models, respectively. Since the ADF method requires gradient information which is unavailable in RF and SVM models, ADF's results are absent in Table \ref{Table-DSS-RF} and Table  \ref{Table-DSS-SVM}. Four main observations are listed below.

\begin{itemize}
	\item {For the TSN, \mytool{} outperforms the baseline approaches significantly, indicating that the size of search space via \mytool{} is much bigger than other approaches. The main reason is the crossover and mutation operators employed in \mytool{}, benefiting an efficient generation of large numbers of new testing samples and avoiding a localized search.}
	
	\item{For the DSN and DSS, \mytool{} can generate about 4 times the discriminatory samples compared with ADF. The advantage of \mytool{} is prominent when compared with AEQUITAS and SG. For example, in the Census dataset with \textit{gender} as the protected attribute, \mytool{} requires only 0.03s (indicated by DSS) to generate a discriminatory sample while AEQUITAS and SG require 2s and 4.13s.}
	
	\item{For the SUR, in most datasets, \mytool{} presents the best effectiveness, except for a few cases where SUR value by \mytool{} is slightly smaller than that of SG. For example, on the Bank  dataset, the SUR by \mytool{} is 57.90\% while SUR by SG is 68.63\%. We figure out that this exceptional observation is caused by the time period setting. That is, SUR by \mytool{} has not achieved a stable value within a time of one hour while SG achieves its highest SUR value. After we extended the time to two hours, the SUR of \mytool{} surpasses SG.}
	
	\item{We observe the stability of methods with changes of models. For all the three models, \mytool{} can achieve a relatively more stable performance with a smaller DSS and a bigger SUR than other approaches. The results indicate a better generalizability of \mytool{} on different models. For AEQUITAS, in the Bank dataset using the SVM model, the SUR is only 0.13\%, much lower than those (sometimes 63.16\%) in MLP and RF models. For SG, in the Census Income dataset using the SVM model, there exist DSN values equal to 0.}
\end{itemize}

\begin{figure}[!t]
	\centering
	\includegraphics[scale=0.4]{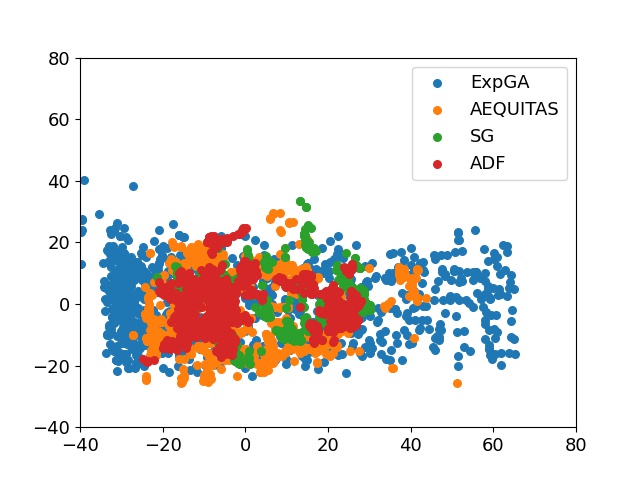}
	\caption{Comparison of the diversity for four approaches.}
	\label{Fig-Sec4-Diversity}
\end{figure}

We also compare the GA to random exploration, keeping seed
exploration the same. The results show that without GA, all the SUR values are not higher than 6\% and the DSS values are 6 times of those using GA, indicating that GA is significant in generating discriminatory samples. The details can be found on our github. 

In summary, even though AEQUITAS is the first approach that proposes a two-phase framework and inspires the following work, its randomly global search scheme leads to detection inefficiency. SG has superior effectiveness than ADF in general and sometimes outperforms \mytool{}. However, SG relies on the heavyweight symbolic execution that requires a substantial time consumption when generating testing samples. ADF appears to be a good choice for white-box deep learning models. However, the gradient information required by ADF is often unavailable in most practical scenarios.

Furthermore, we measure the diversity of generated discriminatory samples for the above four approaches. Specifically, we first randomly select 1,000 discriminatory samples on the MLP model for each approach. Then, we translate their features into two dimensions using PCA. As illustrated in Fig. \ref{Fig-Sec4-Diversity}, we can observe that the coverage area of \mytool{} is larger than others, indicating that the discriminatory samples generated by \mytool{} present the most diversity among the four approaches.

Finally, considering the random factors in the above approaches, we use statistical tests to measure their significance of DSS in terms of two metrics, i.e., p-value and effect size~\cite{arcuri2011practical}. To calculate the p-values, we apply the non-parametric Mann-Whitney-Wilcoxon to compare the DSS of \mytool{} with those of other three baseline approaches. Then, to calculate the effect size metric, we employ the non-parametric Vargha and Delaney's $\widehat{A}_{12}$  statistics.
The results demonstrate that all the p-values are 0, and all the effect sizes are lower than the standard threshold value, i.e., 0.29, indicating that the DSS of \mytool{} is significantly smaller than the others.

\begin{tcolorbox}
	\textbf{Answering RQ~1:} Compared with three baseline approaches, our \mytool{} presents the best efficiency and effectiveness in detecting discriminatory samples from tabular datasets. On average, \mytool{} requires less than 0.2s (indicated by DSS) to detect a discriminatory sample with about 49\% (indicated by SUR) success rate. Moreover, the performance of \mytool{} on diverse models is more stable than those of other approaches, indicating that \mytool{} is model-agnostic and can effectively handle black-box models.    
\end{tcolorbox}

\begin{table}[!t]
	
	\newcommand{\tabincell}[2]{\begin{tabular}{@{}#1@{}}#2\end{tabular}}
	\caption{The performance of \mytool{} using CNN model.}
	\label{Table-Text-CNN}
	\centering
	\scalebox{0.95}{
		\centering
		\begin{tabular}{ll|rrrr} 
			\toprule[1.5pt]
			\multicolumn{2}{c|}{\multirow{2}{*}{Dataset}} & \multicolumn{4}{c}{ExpGA}        \\
			\multicolumn{2}{l|}{}                         & TSN    & DSN   & DSS   & SUR     \\ 
			\hline
			\multirow{2}{*}{IMDB} & gender                & 44,381 & 2,477 & 16.42 & 5.60\%  \\
			& country               & 24,405 & 357   & 118.8 & 1.46\%  \\ 
			\hline
			\multirow{2}{*}{SST}  & gender                & 4,579  & 105   & 14.6  & 2.29\%  \\
			& country               & 4,669  & 252   & 4.8   & 5.40\%  \\
			\bottomrule[1.5pt]
		\end{tabular}
	}
\end{table}

\begin{table}[!t]
	\newcommand{\tabincell}[2]{\begin{tabular}{@{}#1@{}}#2\end{tabular}}
	\caption{The performance of \mytool{} using LR model.}
	\label{Table-Text-LR}
	\centering
	\scalebox{0.95}{
		\centering
		\begin{tabular}{ll|rrrr}
			\toprule[1.5pt]
			\multicolumn{2}{c|}{\multirow{2}{*}{Dataset}} & \multicolumn{4}{c}{ExpGA}        \\
			\multicolumn{2}{l|}{}                         & TSN    & DSN   & DSS   & SUR     \\ 
			\hline
			\multirow{2}{*}{IMDB} & gender                & 77,898 & 1,292 & 13.29 & 1.66\%  \\
			& country               & 17,245 & 257   & 46.76 & 1.49\%  \\ 
			\hline
			\multirow{2}{*}{SST}  & gender                & 10,977 & 620   & 2.39  & 5.65\%  \\
			& country               & 4,932  & 234   & 4.90  & 4.74\%  \\
			\bottomrule[1.5pt]
		\end{tabular}
	}
\end{table}

\begin{table*}[!t]
	\newcommand{\tabincell}[2]{\begin{tabular}{@{}#1@{}}#2\end{tabular}}
	\caption{Comparison results of detecting discriminatory samples between \mytool{} and MT-NLP on IMDB dataset.}
	\label{Table-Text-Compare}
	\centering
	\scalebox{0.95}{
		\centering
		\begin{tabular}{l|l|rrrr|rrrr} 
			\toprule[1.5pt]
			\multirow{2}{*}{Dataset}     & \multirow{2}{*}{Model} & \multicolumn{4}{c|}{ExpGA}      & \multicolumn{4}{c}{MT-NLP}        \\
			&                        & TSN    & DSN   & DSS   & SUR    & TSN     & DSN   & DSS   & SUR     \\ 
			\hline
			\multirow{2}{*}{IMDB gender} & CNN                    & 44,381 & \textbf{2,477} & \textbf{16.42} & \textbf{5.60\%} & \textbf{174,183} & 1,111 & 90.73 & 0.64\%  \\
			& LR                     & 77,898 & \textbf{1,292} & \textbf{13.29} & \textbf{1.66\%} & \textbf{174,183} & 1,322 & 76.25 & 0.76\%  \\
			\bottomrule[1.5pt]
		\end{tabular}
	}
\end{table*}

\begin{table*}[!t]
	\newcommand{\tabincell}[2]{\begin{tabular}{@{}#1@{}}#2\end{tabular}}
	\caption{Comparison results of detecting discriminatory samples before and after retraining.}
	\label{Table-Retraining}
	\centering
	\scalebox{0.95}{
		\centering
		\begin{tabular}{ll|c|rr|cr|rrr|rrr} 
			\toprule[1.5pt]
			\multicolumn{2}{c|}{\multirow{2}{*}{Dataset}} & Sample Added & \multicolumn{2}{l|}{\tabincell{c}{Normal Sample\\ Testing Accuracy}} & \multicolumn{2}{l|}{\tabincell{c}{Discriminatory Sample\\ Testing Percentage}} & \multicolumn{3}{c|}{DSS}     & \multicolumn{3}{c}{SUR}          \\
			\multicolumn{2}{l|}{}                         &              & Before & After                                      & Before & After                                                & Before & After & Impr & Before  & After   & Impr  \\ 
			\hline
			\multirow{3}{*}{Cencus} & gender             & 10\%         &   84.65\%     & 84.41\%                                     & 100\%  & 0.1\%                                                & 0.03   &   0.37    &  12.3x           & 28.70\% & 11.14\% &  61.12\%      \\
			& age                & 10\%         &   84.65\%     & 83.30\%                                     & 100\%  & 3.9\%                                                & 0.12   &    0.23   &   1.9x          & 67.79\% & 19.87\% &  70.69\%   \\
			& race               & 10\%         &    84.65\%    & 84.60\%                                     & 100\%  & 1.7\%                                                & 0.12   &  0.17     &    1.4x         & 33.41\% & 22.50\% &  32.65\%     \\ 
			\hline
			\multirow{2}{*}{Credit}  & gender             & 10\%         &    74.17\%    & 74.22\%                                     & 100\%  & 2.3\%                                                & 0.06   &    0.18   &   3.0x          & 23.12\% &    13.97\%     &   39.57\%           \\
			& age                & 10\%         &   74.17\%      & 75.81\%                                     & 100\%  & 7.8\%                                                & 0.11   &   0.15    &   1.4x          & 73.44\% &  53.62\%       &   26.98\%           \\ 
			\hline
			Bank                     & age                & 10\%         &    89.52\%    & 89.40\%                                     & 100\%  & 8.4\%                                                & 0.13   &   0.33    &  2.5x           & 57.90\% &     41.65\%    &        28.07\%      \\ 
			\hline
			\multirow{2}{*}{IMDB}    & gender             &     2.58\%         & 87.55\%   & 86.24\%                                       & 100\%  & 0.67\%                                               & 16.42  &   211.48    &   12.8x          & 5.60\%  & 0.48\%  &  91.43\%     \\
			& country            &     0.72\%         & 87.55\%   & 85.70\%                                       & 100\%  & 0\%                                                  & 118.8  &   334.07    &    2.8x         & 1.46\%  & 0.45\%  &  69.18\%     \\ 
			\hline
			\multirow{2}{*}{SST}     & gender             &      7.31\%        & 79.50\%   & 78.82\%                                       & 100\%  & 0\%                                                  & 14.6   &   107.72    &  7.4x           & 2.29\%  & 1.16\%  &   49.35\%   \\
			& country            &     1.82\%         & 79.50\%   & 77.70\%                                       & 100\%  & 5.7\%                                                & 4.8    &   576.56    &    120x         & 5.40\%  & 0.31\%  &   94.26\%    \\ 
			\bottomrule[1.5pt]
		\end{tabular}
	}
\end{table*}

\subsection{Performance on Text Datasets}
\label{subsec_text}


To answer RQ2, we randomly select 1,000 samples from the original datasets as our input. Note that we set the iteration parameter for the text datasets $l$ as 20 to stop our algorithm. 

Table \ref{Table-Text-CNN} and Table \ref{Table-Text-LR} list the discriminatory sample detection results by \mytool{} using CNN and LR models, respectively. From the results, we can draw two main conclusions. 

\begin{itemize}
	\item {The results show that \mytool{} can effectively detect discriminatory samples. However, the performance measured by DSS and SUR on text datasets is worse than that on tabular datasets shown in RQ1. The performance difference is due to that the word replacement in \mytool{} employs different strategies for text and tabular datasets. On tabular datasets, a word can be replaced by many candidates. On text datasets,  a word only can be finitely replaced by its synonyms to retain the syntax and semantic correctness of a sentence after word substitution. As a result, DSN calculated on text datasets is smaller than that on tabular datasets.}
	\item {We observe that DSS values on the IMDB dataset are bigger than those on the SST dataset. It indicates that the detection on the former dataset requires a greater time consumption than that on the latter. The average length of samples in the IMDB dataset is ten times larger than that in the SST dataset. The longer the samples, the larger the detection space, leading to a higher time cost in every iteration.}
\end{itemize}

Moreover, we compare the performance of our approach with MT-NLP based on CNN and LR models. Since the source code of MT-NLP is unavailable,  we directly use the performance results evaluated by the work of ~\cite{ma2020metamorphic}. 
Table \ref{Table-Text-Compare} presents the comparison results on IMDB dataset with \textit{gender} protected attribute. Although \mytool{} shows smaller TSN than MT-NLP, its DSS value is about five times smaller than MT-NLP, and the SUR value is at least twice larger than MT-NLP. 


\begin{tcolorbox}
	\textbf{Answering RQ~2:} \mytool{} outperforms the MT-NLP baseline, with at least five times the efficiency and twice the effectiveness in finding discriminatory samples on text datasets.  
\end{tcolorbox}

\subsection{Fairness Improvement through Retraining}
\label{subsec_retraining}

To answer RQ3, we employ a data augmented method, used in AEQUITAS~\cite{udeshi2018automated} and ADF~\cite{zhang2020white}, to investigate the model fairness improvement imposed by the discriminatory samples detected by \mytool{}. Following the data augmented method, given a target model, we execute \mytool{} to generate discriminatory samples from a dataset. A portion of these discriminatory samples are retained in the original dataset, constructing a new dataset. Based on the updated dataset, we retrain the model and re-execute our \mytool{} to test the model fairness. 

To verify the fairness of retrained model, we randomly select 1,000 discriminatory samples as the ground-truth. Since RQ2 shows that the DSN on text datasets is much smaller than that of tabular datasets, we use half of the generated discriminatory samples for retraining and the remaining for testing.

The \textit{Dataset} column and \textit{Sample Added} column in Table \ref{Table-Retraining} list
the datasets with the protected attributes, and the percentage of generated discriminatory samples that remained for constructing new datasets in this experiment. On tabular datasets, $10\% \times |original ~dataset|$ discriminatory samples are reserved for retraining. On text datasets, the percentages are calculated based on the size of generated discriminatory samples. 

The Columns 3-12 in Table \ref{Table-Retraining} show experiment results, i.e., the model accuracy and performance of \mytool{} before and after retraining models. 
\textit{Normal Sample Testing Accuracy} indicates the model accuracy calculated against normal samples in the datasets, and \textit{Discriminatory Sample Testing Percentage} is the corresponding accuracy against the ground-truth discriminatory samples. DSS and SUR show the efficiency and effectiveness of \mytool{}. \textit{Impr} columns denote the DSS (or SUR) improvements of \textit{After} DSS (SUR) vs. \textit{Before} DSS (SUR). These measurements show that:
\begin{itemize}
	\item {The model accuracies of detecting normal samples are nearly unchanged after retraining models on datasets augmented with the discriminatory samples generated by \mytool{}. This observation indicates that the data augmented method has little influence on the classification performance of normal samples.}

	\item {After retraining, on average, more than 97\% of the original discriminatory samples are not biased to new models, indicating that  model fairness has been enhanced significantly. }
	
	\item {The \textit{Impr} of DSS presents that a much longer time is required to detect a discriminatory sample by \mytool{} for re-trained models. For example, in SST dataset using \textit{country} as the protected attribute, after searching about three hours, only 18 discriminatory samples are detected.}
	
	\item {The SUR values drop by an average of 56\% (up to 94.26\% in SST dataset), indicating the greater difficulty for a successful generation of a discriminatory sample.}
\end{itemize}

\begin{tcolorbox}
	\textbf{Answering RQ~3:} After retraining models via a data augmented method using discriminatory samples generated by our \mytool{},  the model fairness can be considerably improved, i.e., more than 97\% of the original discriminatory samples are not biased to new models. Furthermore, the efficiency and effectiveness of generating discriminatory samples on new models become much lower compared with the original models while keeping the testing accuracies on normal samples unaltered. 
\end{tcolorbox}

\section{Threats to Validity}
\label{sec_threats}

\subsection{Threats to Internal Validity}
\noindent\textbf{Interpretable Methods}: Our \mytool{} relies on  interpretable methods, such as LIME~\cite{ribeiro2016should} and SHAP~\cite{lundberg2017unified}, to obtain explanation results for an initial seed selection. In practice, it is difficult for interpretable methods to guarantee perfect explanation results. Consequently, the probably missing or mismatching sensitive features might affect our seed sample quality. To mitigate this threat, our future work will explore more interpretable methods.

\noindent\textbf{Retraining to Improve Model Fairness}: In RQ3, we test the model fairness using a data augmented method by imposing discriminatory samples detected by \mytool{}. Although the model fairness has been improved a lot, we are not claiming that \mytool{} radically combs out the model fairness issue. In the future, a considering of the fairness characteristic during data preparation and model training should be a promising exploration direction~\cite{d2017conscientious}. 

\subsection{Threats to External Validity}

\noindent\textbf{Limited Experiment Datasets}: We evaluate \mytool{} on five benchmark datasets that are used by baseline approaches. The results demonstrated that \mytool{} outperforms baselines on diverse benchmarks, though we cannot conclude that \mytool{} would achieve a good performance on other datasets with other protected attributes. Once new datasets are available, we will further verify \mytool{}.

 \noindent\textbf{Construction of Knowledge Graph}: We construct a prior knowledge graph to identify the words related to protected attributes. Currently, this graph only involves a limited number of protected attributes but sufficient for our experimental evaluation. When using \mytool{} to test model fairness regarding other protected attributes, the knowledge graph should be extended to express more domains.
\section{Related Work}
\label{sec_relatedWork}


\noindent\textbf{Fairness Testing}: Recently, software fairness testing has received much attention. Galhotra \textit{et al.}~\cite{galhotra2017fairness} pioneered a fairness test method named THEMIS. THEMIS distinguishes the group discrimination and individual discrimination for software fairness testing.  However, the strategy of randomly generating test cases in THEMIS is inefficient. Following the definitions coined by THEMIS, Udeshi \textit{et al.}~\cite{udeshi2018automated} proposed AEQUITAS, achieving a better efficiency by a directional searching scheme. This search scheme makes it difficult for AEQUITAS to cover samples in diverse distributions.  Agarwal \textit{et al.}~\cite{aggarwal2019black} proposed SG based on program symbolic execution. SG uses LIME to generate local perturbation samples to build a decision tree and then analyzes each tree path to generate test inputs. Similar to SG, our \mytool{} also leverages the interpretable method LIME. The difference is that \mytool{} uses explanation results to identify high-quality seed samples. ADF~\cite{zhang2020white} is a detection method specifically designed for DNN networks. ADF requires gradient information from the model architecture, thus restricting its practical usage. Chakraborty \textit{et. al}~\cite{chakraborty2021bias} 
thought that data imbalance and improper data label are two main reasons for model biased, thus they proposed Fair-SMOTE to achieve group fairness. Fair-SMOTE first synthetically generates new data points for all the subgroups except the subgroup having the maximum number of data points to solve data imbalance in training data. Then, it find out and remove biased samples in training data to eliminate biased labels. It is worth noting that our work focuses on individual fairness, while Fair-SMOTE focuses on group fairness. Therefore, we do not compare our work with Fair-SMOTE in this paper.

In summary, compared with these approaches, our \mytool{} is a lightweight model-agnostic individual fairness testing method, able to handle diverse and large-scale scenarios.


\noindent\textbf{Interpretable Methods}: This work is also related to researches related to machine learning interpretable methods. Ribeiro \textit{et al.} proposed a fast and effective method LIME~\cite{ribeiro2016should} that fits a local decision boundary with a simple linear regression model. The weight of the linear model represents the importance of features. Lundberg and Lee~\cite{lundberg2017unified} proposed SHAP, generates a set of shap values based on linear functions to represent the coefficients of each feature. Unlike LIME, SHAP defines multiple SHAP kernels as a weight function. At the same time, inspired by game theory, SHAP defines three properties to constrain the fitting process.
Guidotti \textit{et al.}~\cite{guidotti2018local} proposed a rule-based black box model interpretable method called LORE, which  first uses genetic algorithms to generate two sample sets with completely different labels. Then, it constructs a decision tree and extracts a set of rules from the tree as a local explanation.
Considering that a single linear function cannot fit the highly nonlinear decision boundary well, Guo \textit{et al.}~\cite{guo2018lemna} used multiple linear models to approximate the local decision boundary and chose the one with the highest accuracy as interpretation result.
 




\section{Conclusion}
\label{sec_conclusion}
This work proposes \mytool{}, an explanation-guided method through the GA for software fairness testing. The novelty of \mytool{} is the combination of explanation results and GA, thus determining its high efficiency and effectiveness in discriminatory sample detection. \mytool{} is also model-agnostic and can handle black-box models in diverse scenarios. The evaluation experiments demonstrate that \mytool{} can detect discriminatory samples much faster with a higher success rate than four state-of-the-art methods, both on the text and tabular benchmarks. Augmented with the discriminatory samples generated by \mytool{}, the fairness of tested models has a substantial improvement through retraining.

\section*{Acknowledgment}
This work was supported by National Key R\&D Program
of China (2018YFB1004500), National Natural Science Foundation of China (61902306, 62002280, 61632015, 61602369, U1766215, 61772408, 61702414, 61833015), China Postdoctoral Science Foundation(2019TQ0251, 2020M673439, 2020M683507), Innovative Research Group of the National Natural Science Foundation of China (61721002), Ministry of Education Innovation Research Team (IRT\_17R86), Youth Talent Support Plan of Xi'an Association for Science and Technology (095920201303).

\bibliographystyle{ACM-Reference-Format}
\bibliography{sample-base}

\end{document}